\documentclass[10pt,twocolumn,letterpaper]{article}

\usepackage{cvpr}
\usepackage{times}
\usepackage{epsfig}
\usepackage{graphicx}
\usepackage{amsmath}
\usepackage{amssymb}
\usepackage{subfig}
\usepackage{float}

\usepackage[breaklinks=true,bookmarks=false]{hyperref}

\cvprfinalcopy 

\begin{document}

\title{\vspace{-1.25cm}Architectural Resilience to Foreground-and-Background Adversarial Noise}
\author{Carl Cheng\\
\tt carl@caltech.edu\\
{\tt carlc@xsense.ai}
\and
Evan Hu\\
{\tt evanhu@berkeley.edu}
}

\maketitle

\begin{abstract}
   Adversarial attacks in the form of imperceptible perturbations of normal images have been extensively studied, and for every new defense methodology created, multiple adversarial attacks are found to counteract it. In particular, a popular style of attack, exemplified in recent years by DeepFool and Carlini-Wagner, relies solely on white-box scenarios in which full access to the predictive model and its weights are required. In this work, we instead propose distinct model-agnostic benchmark perturbations of images in order to investigate the resilience and robustness of different network architectures. Results empirically determine that increasing depth within most types of Convolutional Neural Networks typically improves model resilience towards general attacks, with improvement steadily decreasing as the model becomes deeper. Additionally, we find that a notable difference in adversarial robustness exists between residual architectures with skip connections and non-residual architectures of similar complexity. Our findings provide direction for future understanding of residual connections and depth on network robustness. 
\end{abstract}

\section{Introduction}

Convolutional Neural Networks (CNNs) are a network of neurons that utilize convolutions rather than the traditional matrix multiplication. Certain groups of nodes and neurons can be simplified into  "layers", which are specified based on function and structure, and every layer must be able to receive a weighted input and transform said input into an output. CNNs are composed primarily of two types of layers: convolutional and pooling layers. Convolutional layers can undergo n-dimensional convolutions (2 dimensions is shown):
\begin{equation}
    (f\ast g)(c_1, c_2) = \sum_{a_1, a_2} f(a_1, a_2) \cdot g(c_1-a_1,~ c_2-a_2)
\end{equation}
which, in turn, allow for backpropagation, allowing the CNN to retrace its steps in-case it ever approaches an incorrect solution\cite{lecun1989backpropagation}. On the other hand, pooling layers allow for feature detection, using spatial maps to group nearby and like pixels. In conjunction with other types of layers and in distinct arrangements, certain architectural styles have risen to the top, becoming state-of-the-art classifiers.

These recent advancements in machine learning have led to "Deep learning", an extension dealing with deeper neural networks (DNNs), particularly Deep CNNs. For these models, image classification remains one of the most popular and robust tasks\cite{zha2015exploiting, lee2017going}, tasking the DNN with recognizing patterns in computer vision and speech. Classification of images stands frequently as a benchmark for newly developed architectures and data augmentation methods\cite{yu2016automated, simonyan2014very, huang2019gpipe, gong2020maxup}. 

\begin{figure}[t]\vspace{-3mm}
{\centering
\begin{tabular}{c}\vspace{-3mm}
\subfloat{\includegraphics[width=.64\linewidth]{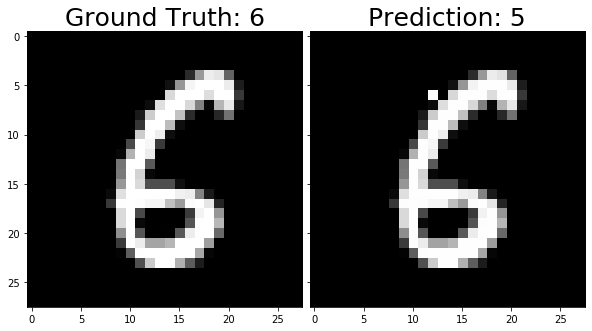}}\\
\subfloat{\includegraphics[width=.64\linewidth]{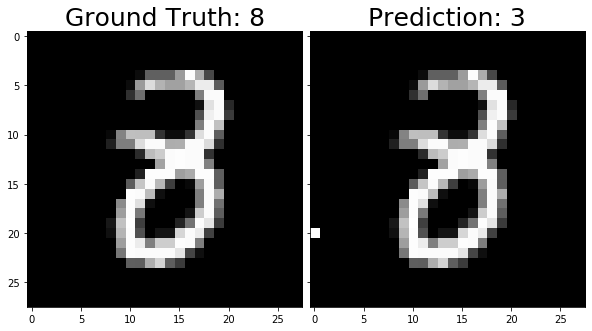}} 
\end{tabular}
   \vspace{-3mm}\caption{(Left) Images from MNIST that model ResNet-50 initially classified correctly. (Right) Images from MNIST, now perturbed, classified incorrectly.}
}
\end{figure}

However, only within the past decade did hardware improvements even allow for DNNs, which now handle its massive amounts of training data on GPUs. Deep neural networks have come a long way, from tweaking architecture topology\cite{larsson2016fractalnet}  to changing from "deep" to "wide" \cite{zagoruyko2016wide}. Though many NNs exist, we derive our focus from select deep networks and from a fully-connected, non-convolutional network. Specific model and learner information is further explained in Section 3.

Regardless of the model, each should be able to train effectively enough so that it can generalize a variety of situations. For example, a cat classifier ideally should be able to label a striped cat, a spotted cat, and a tabby cat all as a "cat". Frequently, however, the model will not be able to generalize a variety of situations that human comprehension can easily determine. Most DNNs are widely susceptible to confidently misclassifying images using perturbations nearly imperceptible to the human eye (see Figures 1 and 2), which is counter-intuitive to conventional human uncertainty.


\begin{figure}[t]\vspace{-4mm}
{\centering
\captionsetup{singlelinecheck=off}
\begin{tabular}{ccc}
\subfloat{\includegraphics[width=0.32\linewidth]{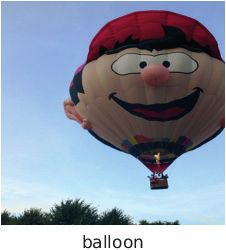}} \subfloat{\includegraphics[width=0.32\linewidth]{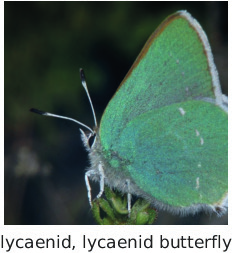}} \subfloat{\includegraphics[width=0.32\linewidth]{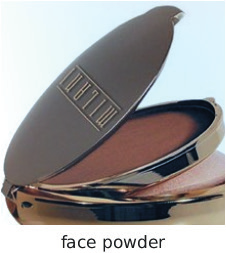}}\\
\subfloat{\includegraphics[width=0.32\linewidth]{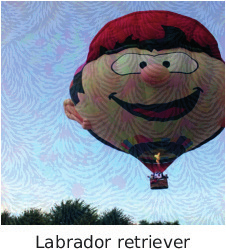}} \subfloat{\includegraphics[width=0.32\linewidth]{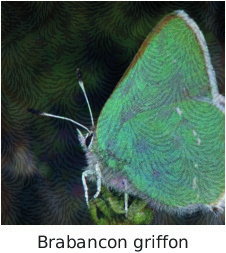}} \subfloat{\includegraphics[width=0.32\linewidth]{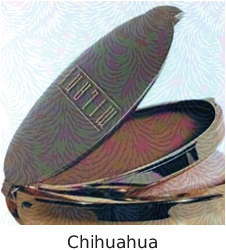}}
\end{tabular}
\vspace{-2mm}\caption[]{Row 1 contains natural images in the ImageNet dataset. Row 2 contains the original images, now altered, given nearly imperceptible perturbations\cite{moosavi2017universal}. These perturbations are universal; if $\rho$ denotes the distribution of vectors of images, $\hat{k}$ defines a classification function, and $v$ is the perturbation vector, they are of criteria such that:
\begin{align}\tag{2}
\hat{k}(x + v) \neq \hat{k}&(x) \text{ for "most" } x \sim \rho
\end{align}}
}
\end{figure}

But just how resistant are these DNNs to extremely general attacks? Imperceptible perturbations are no longer "imperceptible" on datasets consisting of small-sized images (e.g. 28x28 Fashion-MNIST\cite{xiao2017fashion}), and black-box and white-box attacks (although possible on small-image datasets) only aim to give a network an especially specific attack capable of tricking the model into outputting a unique and incorrect label\cite{papernot2017practical}. Our goal isn't to force an autonomous vehicle to classify a stop sign as a yield sign. Rather, we propose four new general perturbations, each capable of being executed on any dataset. By executing each perturbation on each model in varying degrees, we are able to visualize the differences between our attack methodologies and architecture resilience. Our noise generations are then compared to attacks of specialized nature, for example DeepFool and Boundary Attacks.

In this paper, we discover various characteristics of popular model architectures using comparisons of learner results, both within and between our various perturbation methods. Architecture features and depth are investigated via comparing how they react to each magnitude of each attack. Insight is also gained on a model's response towards perturbations in general, and we see if classification accuracy curves follow a certain shape.

\section{Related Work}

Despite their incredible success in the field of image classification amongst many others, it is well known that DNNs can be extremely vulnerable to adversarial attacks: carefully crafted modifications to inputs that can cause models to misclassify. Since it was first shown that attacks of an imperceptible nature could could be successful\cite{szegedy2013intriguing, goodfellow2014generative}, multitudes of new attack methodologies have been proposed since then. 

More recently, DeepFool and Carlini-Wagner attacks\cite{carlini2016evaluating} have been proven highly effective in bypassing existing adversarial defences and substantially reducing their effectiveness. However, these two, along with the majority of successful and notable attacks, require a white-box scenario in which full access to the classifier and its capabilities is required. As such, inherent limitations exist to the practical application of these attacks as such access is not always so easily found.

Other researchers have delved into the concept of transferable adversarial attacks, which can find success in causing misclassification across a broad range of model architectures. \cite{papernot2016transferability} manages to exploit this property of transferability by constructing substitutes of an unknown black-box model and creating adversarial attacks against the substitute which are also successful on the black-box model. Further studies on transferability have also lead to established strategies for the generation of large proportions of transferable examples\cite{liu2016delving}.

Despite many defences proposed for every successful attack (notably MagNet\cite{magnet} and "Efficient Defenses"\cite{efficientdefenses}) it has also been shown that such defences can be easily fooled as well\cite{carlini2017adversarial}. Moreover, the inevitability of adversarial examples has also been discussed\cite{shafahi2018adversarial}, finding that for many classes of problems, adversarial examples are inescapable. 

Instead of attempting to develop effective defenses for any specific attack, we build upon previous work in establishing network architectures and topologies that are more robust to adversarial examples\cite{gu2014deep}. We seek to provide guidelines on promoting resilience in networks through an in depth examination of various architectures and general attacks.

\section{Methods}

Although not necessarily a true "attack", we will define each pixel generated using our adversarial noise techniques as a basic adversarial attack. One generation results in one pixel changed, thus, one attack was executed.

\subsection{Dataset}

MNIST, a dataset of 70,000 28x28 pixel hand-written digits, was used to train each model. We utilize a single-subject dataset (a singular foreground and a singular background) to minimize the complexity of the task and generalize our experiments for more scenarios. For example, using a dataset of traffic lights may only be useful in an autonomous driving context or in a circle \& color detection context.

All images are resized from 28 by 28 to 56 by 56. Consequently, the "single-pixel" attacks are upscaled to four pixels in a 2 by 2 square. Relative to the original dimensions, the attack on MNIST is preserved.
\subsection{Adversarial Noise Generations}

While there have been various papers documenting the strongest adversarial examples\cite{46561}, few have investigated the threshold at which the model can successfully classify given an image perturbation. Even fewer have considered the weakest attack a DNN can misclassify on, although some graceful methodologies do exist\cite{brendel2017decision}. Alongside the simplest dataset of MNIST, we use the simplest attacks, which we divide into four distinct categories: \textbf{Random Pixel Invert} (control noise), \textbf{White to Black}, \textbf{Black to White}, and \textbf{Edge to Altered}. We start our attacks at a single-pixel generation, and subsequently increase the proportion of pixels affected until the number of attacks reaches 200. By starting with an exceedingly weak one-pixel attack, we can slowly increase the noise magnitude by increasing pixel count. This enables us to capture all degrees of perturbation possible, from the weakest to the strongest.

\textit{For the following methods, $b$ is the brightness of the pixel, of which values range from $0$ (black) to $255$ (white). $c$ is the constant used in the corresponding inversion algorithms. Within the 3D array, $n$ is the image index, and $x$ and $y$ are pixel location indices $x$ and $y$, respectively.}

\paragraph{Random Pixel Noise.} Inverts a random pixel. For pixels with $b > 127$, noise is generated using Pythonic equation:
\begin{equation}\tag{3}
\begin{split}
    &int(\frac{255 - b}{c}) \text{ (by default }c = 1.25\text{) at location }(n, x, y)\\\label{e1}
    & \text{ for all tensors in the MNIST test set.}
\end{split}
\end{equation}
For all other pixels, noise is generated using Pythonic equation:
\begin{equation}\tag{4}
\begin{split}
    &int(255 - \frac{b}{c}) \text{ (by default }c = 1.25\text{) at location }(n, x, y)\\\label{e2}
    & \text{ for all tensors in the MNIST test set.} 
\end{split}
\end{equation}

$c$ is any positive number, including floats. If $c < 1$, after a continuous amount of generations, the image will continually approach gray. If $c > 1$, the image will instead approach black and white extremes. Setting $c = 1.25$ operates as our control noise because the model should approach a more extreme version of the image after each attack.

\paragraph{White Pixel to Black Pixel Noise.} Inverts a random white pixel (defined as $b > 0$) to black using equation \eqref{e1} with $c = 2$. In terms of human vision, this is a foreground-to-background attack.

\vspace{-1mm}\begin{figure}[h]
\centering{
   \includegraphics[width=0.6\linewidth]{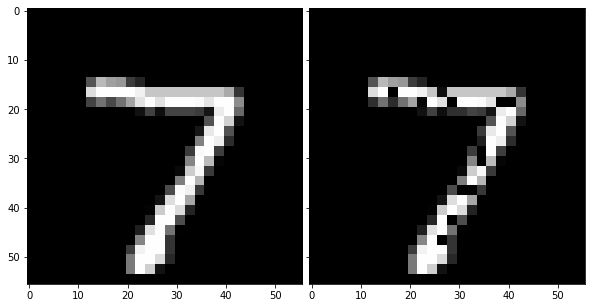}
   \vspace{-4mm}\caption{Example of White to Black attacks, with attacks $= 20$.}
}
\end{figure}

We choose $b > 0$ as we set $b > 100$ previously, but after 20 or more attacks, an outline of the digit appeared in gray pixels ($0 < b < 100$). The model still managed to misclassify frequently when given the gray outline of the digit, even though the number appeared to be obvious.

\paragraph{Black Pixel to White Pixel Noise.} Inverts a random black pixel (defined as $b < 100$) to white using equation \eqref{e2} with $c = 2$. In terms of human vision, this is a background-to-foreground attack.

\vspace{-1mm}\begin{figure}[h]
{\centering
   \includegraphics[width=0.6\linewidth]{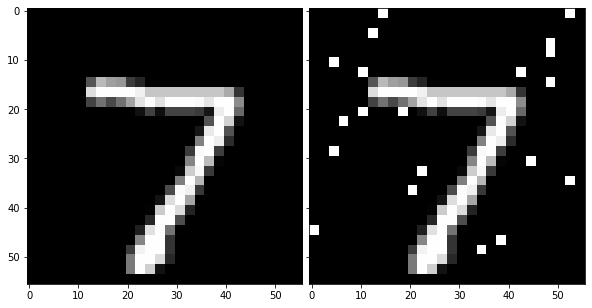}
   \vspace{-4mm}\caption{Example of Black to White attacks, with attacks $= 20$.}
}
\end{figure}
\paragraph{Edge Pixel to Altered Pixel Noise.} Inverts a random edge pixel. An edge pixel is defined as any pixel that is vertically, horizontally, or diagonally adjacent to a pixel of opposite color. The opposite color criteria is met when one pixel has $b > 127$ and the other has $b \le 127$, or vice versa. If the selected edge pixel has $b > 127$, equation \eqref{e1} is used. Otherwise, equation \eqref{e2} is used.

\vspace{-1mm}\begin{figure}[H]
{\centering
   \includegraphics[width=0.6\linewidth]{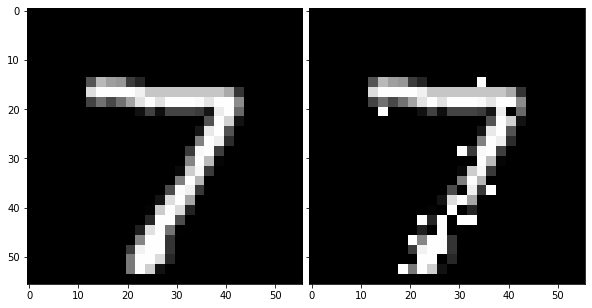}
   \vspace{-4mm}\caption{Example of Edge to Altered attacks, with attacks $= 20$.}
}
\end{figure}

We choose these noise generations based on two factors; human deduction and 2D image characteristics.

By inverting pixels based on certain criteria, none of these generations preserve object identity, meaning that many human perception tools (such as depth and the ventral visual stream\cite{dicarlo2012does, snow2014real}) become inapplicable. We eliminate object-identity-preserving transformations, subsequently placing our generations in the same category as imperceptible perturbations.

But beyond avoiding warp and rotate attacks, we choose these particular perturbations as they simulate various possibilities for noise in 2D images. For example, consider White to Black. This method was made to imitate degradation and fading away of an image foreground over time (similar to how paint on a wall chips and fades away). Likewise, Black to White corresponds to random background noise scattered around, and Edge to Altered corresponds to "bad handwriting"; digits that are written too thin, too thick, or uneven.

\subsection{Learners}

For our custom Fully-Connected (FC) network, we theorized the possibility of splicing pooling layers within and changing the input dimensions after each pooling layer, but after only adding one pooling layer, the result failed to even come close to its original MNIST test-set accuracies. Though it's very possible that a particularly arranged Pooling $\in$ FC architecture may yield favorable results (VGG architectures are more to be considered FC $\in$ Convolutional, see Figure 6), in most cases, splicing pooling layers in an otherwise Fully-Connected network would adversely affect the model and oversimplify the learning process.

Each model is trained on the MNIST train set with a starting learning rate of $1e-2$ and momentum $0.9$. The learning rate decays by a factor of $10$ every epoch. The training process continues up until 99.0\% validation accuracy is achieved or surpassed (the only exception being the Fully-Connected network). Though the number of epochs between each model becomes volatile, each should classify at a very similar correctness. This is further authenticated in Section 4, Table 1, where all models score similarly well on an unaugmented test set.

\begin{figure}[t]
{\centering

   \includegraphics[width=0.87\linewidth\vspace{-1mm}\hspace{-2.4mm}]{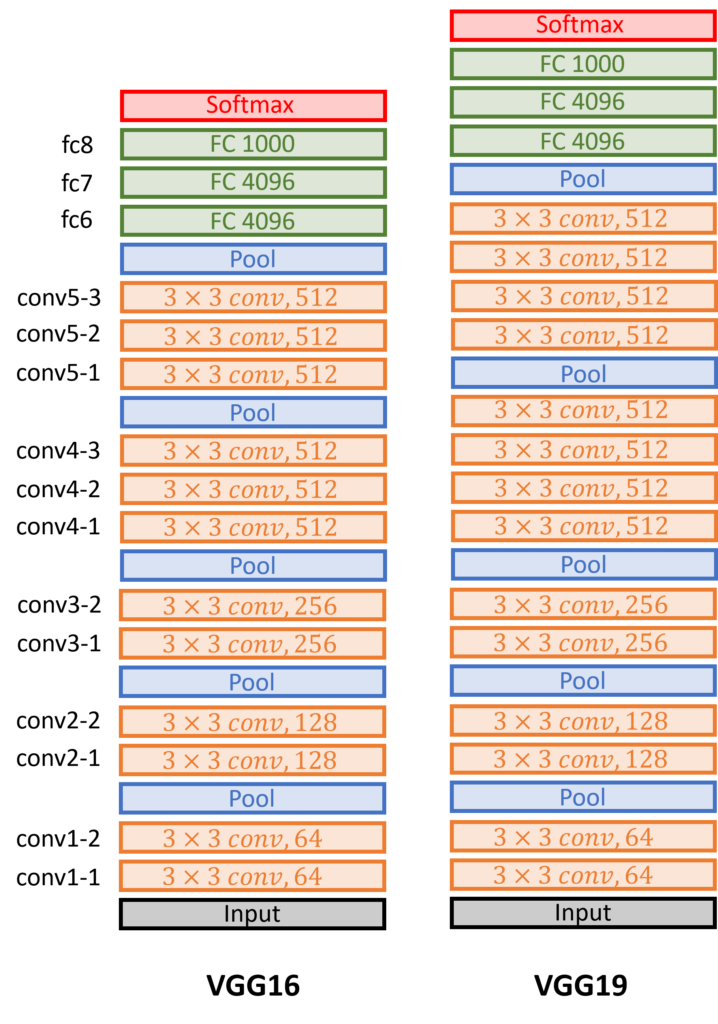}

   \vspace{-3mm}\caption{Example of Fully Connected Layers (FC 4096) placed after convolutional and pooling layers\cite{matic_jordacevic_kalu_2018}. FC 1000 is the output layer. Traditional VGG-16 and VGG-19 nets are shown.}
}
\end{figure}

\paragraph{Fully Connected Network.} A feed forward neural network, consisting of a single 3136-unit hidden layer, is trained using our rules and methods explained above. 3136 was chosen to equal to the number of pixels in our upscaled 56x56 images.

\paragraph{ResNet-50.} A default ResNet-50, consisting of a 50-layer convolutional network with residual connections skipping one or more layers\cite{he2016deep}, is trained using our rules and methods explained above.

\paragraph{ResNet-50V2.} A default ResNet-50V2 is trained using our rules and methods explained above. The difference from the original ResNet-50 exists that a batch normalization layer, then a ReLU function, is placed behind the initial convolutional layer\cite{he2016identity}.

\paragraph{VGG-19.} A default VGG-19, shown in Figure 6, is trained using our rules and methods explained above.

\paragraph{EfficientNet-B1.} A default EfficientNet-B1 is trained using our rules and methods explained above. Rather than scaling up height, width, and image resolution independently, EfficientNet is upscaled via a compound scaling method, consisting of:
\begin{align}\tag{5}
\begin{split}
    \text{depth: }& d = \alpha^\phi\\
    \text{width: }& w = \beta^\phi\\
    \text{res: }& r = \delta^\phi
\end{split}
\end{align}
where $\alpha$, $\beta$, and $\delta$ are constants $\ge 1$ determined by a grid search\cite{tan2019efficientnet}. The architecture consists of a 3x3 Convolutional layer, followed by various mobile inverted bottleneck MBConv layers\cite{tan2019mnasnet, sandler2018mobilenetv2}.

\paragraph{ResNet-101, ResNet-34, and ResNet-18.} Default ResNets of deeper and shallower architectures are used for differing depths within one architecture, ascertaining the effect of depth on attack resilience

\subsection{Experiments}

The models were evaluated on each type of attack, starting with attacks $ = 0$ (the default, unaugmented test set). Attacks were subsequently increased by increments of 10 to a max of 200 attacks, with the evaluation accuracies aggregated in a 5x20 (5 different attacks, 20 iterations) array for each model. After repeating this process 4 times, the resulting accuracies were averaged to produce the final set.

Two separate experiments were conducted through this process: one involving the first 5 learners, omitting depth variant ResNets and keeping only ResNet-50s, and the other including only ResNets with varying depths.

\section{Results}

\subsection{Overall Results}

Displayed in Figure 7 are each model's test accuracy under every attack, given a certain number of attacks. A table of important values can be found on the following page.

\renewcommand{\thefigure}{7}
\begin{figure}[H]
{\centering
\subfloat{\includegraphics[width=1.009\linewidth]{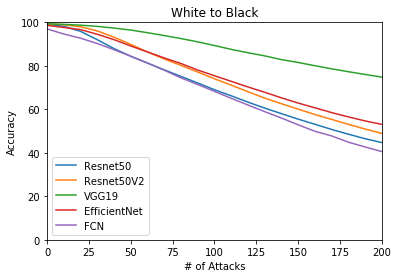}}\\
}
\end{figure}
\begin{figure}\ContinuedFloat
{\centering
\vspace{-4mm}\subfloat{\includegraphics[width=1.009\linewidth]{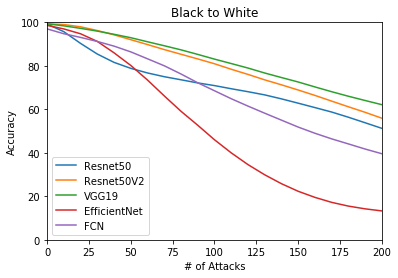}}\\
\vspace{-4mm}\subfloat{\includegraphics[width=1.009\linewidth]{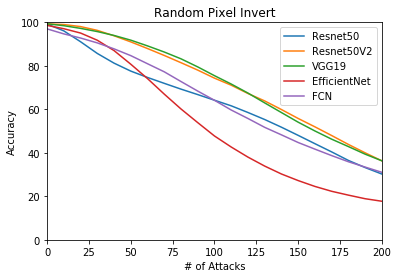}}\\
\vspace{-4mm}\subfloat{\includegraphics[width=1.009\linewidth]{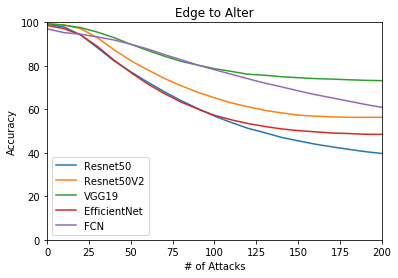}}
\vspace{-2mm}\caption{Average adversarial accuracies (with sample size $n = 5$) for ResNet-50, ResNet-50V2, VGG-19, EfficientNet-B1, and our Fully-Connected Network are plotted from attack $ = 0$ to attack $ = 200$. Similar trends were observed when the sample size was increased.}
}
\end{figure}

\newpage
The results show slight differences in performance between ResNet50 and ResNet50V2, with the latter consistently testing ~5-10\% higher in accuracy on attacks. This may be attributed to the slightly higher accuracy of ResNet-50V2 in training, but it is unlikely that this difference is the sole factor.

Interestingly enough, VGG-19 seems to be the best performer in nearly all regards, with the exception of our control attack. EfficientNet-B1 displayed poor robustness to our attack methods, with it ranking in or near the bottom for three out of our four attacks. EfficientNet also performed significantly worse on Black to White and Random Pixel Invert attacks, testing between 20-30\% lower accuracy than the next lowest performer. 

We included a Fully Connected Network model as a control, though it performed surprisingly well in contrast to our expectations. Although it had the lowest accuracies in the White to Black attacks, it outranked EfficientNet in every other attack and even achieved second best accuracy consistently throughout the Edge to Alter attacks. It's relative success given its simplicity, along with VGG19, suggest a strong direction for understanding the differences between residual and non residual architectures in networks.

The structure of each curve is presumably related to a combination of model architecture and attack type; looking at EfficientNet's and ResNet's curves, each seems to be somewhat sigmoidal, but ResNet suggests an inverted sigmoid in Random Pixel Invert and Black to White (whereas EfficientNet remains a consistent sigmoid). And considering that the FCN typically only presents linear curves, it becomes evident that different architectures respond in fairly different ways depending on the attack received.

\begin{table}[h]
{\centering
\begin{tabular}{|l|c|c|c|c|}
\hline
    Attacks & WtB & BtW & RPI & EtA\\
\hline
    \multicolumn{5}{|c|}{ResNet-50} \\
\hline
\hline
0& 99.20&	99.20&	99.20&	99.20\\
\hline
10& 98.17&	95.72&	96.08&	97.90\\
\hline
40& 87.93&	81.59&	81.25&	82.31\\
\hline
100& 69.01&	70.92&	64.19&	56.90\\
\hline
200& 44.71&	51.25&	30.24&	39.70\\
\hline
\hline
\multicolumn{5}{|c|}{ResNet-50V2}\\
\hline
0& 99.30&	99.30&	99.30&	99.30\\
\hline
10& 98.91&	99.00&	99.00&	98.85\\
\hline
40& 93.19&	94.26&	93.83&	87.38\\
\hline
100& 74.05&	80.96&	74.35&	65.31\\
\hline
200& 48.91&	55.91&	36.31&	56.34\\
\hline
\hline
\multicolumn{5}{|c|}{VGG19}\\
\hline
0& 99.31&	99.31&	99.31&	99.31\\
\hline
10& 99.07&	98.40&	98.47&	98.70\\
\hline
40& 97.36&	94.51&	93.98&	92.98\\
\hline
100& 89.35&	83.10&	75.48&	78.71\\
\hline
200& 74.83&	62.14&	36.33&	73.19\\
\hline
\hline
\multicolumn{5}{|c|}{EfficientNet-B1}\\
\hline
0& 98.70&	98.70&	98.70&	98.70\\
\hline
10& 97.65&	96.98&	97.08&	97.14\\
\hline
40& 92.05&	85.99&	87.02&	82.63\\
\hline
100& 75.44&	46.02&	47.75&	57.24\\
\hline
200& 53.07&	13.28&	17.73&	48.49\\
\hline
\hline
\multicolumn{5}{|c|}{Fully Connected Network}\\
\hline
0& 97.12&	97.12&	97.12&	97.12\\
\hline
10& 94.64&	94.76&	94.70&	95.30\\
\hline
40& 87.41&	89.07&	87.82&	91.88\\
\hline
100& 68.27&	68.55&	64.16&	78.23\\
\hline
200& 40.57&	39.58&	31.04&	60.90\\
\hline
    \multicolumn{5}{@{}p{4.0cm}}{\footnotesize WtB $=$ White to Black}\\
    \multicolumn{5}{@{}p{4.0cm}}{\footnotesize BtW $=$ Black to White}\\  
    \multicolumn{5}{@{}p{4.0cm}}{\footnotesize RPI $=$ Random Pixel Invert}\\  
    \multicolumn{5}{@{}p{4.0cm}}{\footnotesize EtA $=$ Edge to Altered}  
\end{tabular}
\vspace{-2mm}\caption{Overall Attack Accuracies. Important values are shown.}
}
\end{table}

\subsection{Results Compared to Other Attacks}

\begin{table}[H]
{\centering
\begin{tabular}{|l|c|}
\hline
\multicolumn{2}{|c|}{Attacks Classified by ResNet-50/CNN}\\
\hline
\hline
DeepFool& 13.0\\
\hline
Boundary& 21.0\\
\hline
WtB $=$ (200)& 44.7\\
\hline
BtW $=$ (200)& 51.3\\
\hline
EtA $=$ (200)& 39.7\\
\hline\hline
Pointwise& 91.0\\
\hline
WtB $=$ (40)& 87.9\\
\hline
BtW $=$ (40)& 81.6\\
\hline
EtA $=$ (40)& 82.3\\
\hline
\multicolumn{2}{@{}p{4.0cm}}{\footnotesize ( ) $=$ Number of Attacks} 
\end{tabular}
\caption{Comparative results for our adversarial generations with DeepFool, Boundary, and Pointwise attacks. DeepFool and Boundary typically alter $\ge200$ pixels in the image, while Pointwise typically alters $\approx40$ pixels.}
}
\end{table}

DeepFool\cite{moosavi2016deepfool}, Boundary Attacks\cite{brendel2017decision}, and Pointwise Attacks were performed on the test set and classified by a CNN with 99.1\% MNIST test set accuracy\cite{schott2018towards}, similar (both in structure and test set accuracy) to ResNet-50. We thereby compare these accuracies with our noise generation accuracies for ResNet-50.
    
The "degree" of attack is determined by the number of pixels altered; equal degrees of comparison will be used, meaning that the compared attacks will affect equal proportions of their respective images.

Based on these comparisons, our attacks resemble Pointwise attacks more closely compared to DeepFool and Boundary perturbations, of which Pointwise represents the weakest extreme and the latter two represent the strongest extreme.

\newpage

\subsection{ResNet Depth Results}

We choose depth as our variable and display a graph of ResNet at various depths, each classifying on every attack, given a certain number of attacks.

\renewcommand{\thefigure}{8}
\begin{figure}[h]
{\centering
\vspace{-1.4mm}\subfloat{\includegraphics[width=1.005\linewidth]{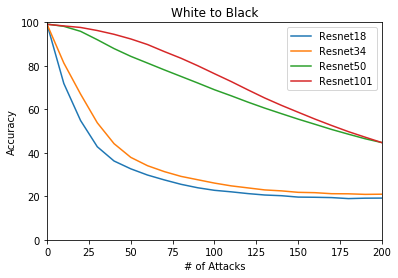}}\\ \vspace{-4mm}\subfloat{\includegraphics[width=1.005\linewidth]{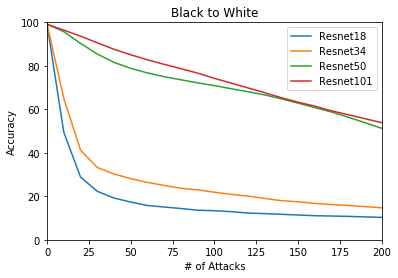}}\\
\vspace{-4mm}\subfloat{\includegraphics[width=1.005\linewidth]{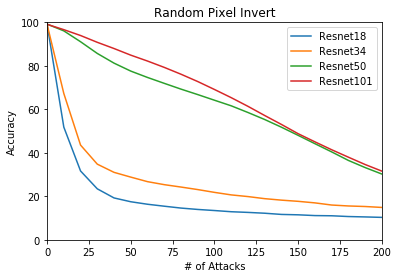}}\\
}
\end{figure}
\begin{figure}\ContinuedFloat
{\centering
\vspace{-4mm}\subfloat{\includegraphics[width=1.005\linewidth]{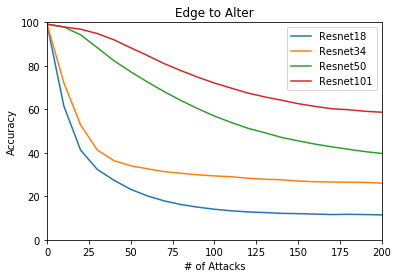}}
\caption{Average adversarial accuracies (with sample size $n = 5$) for ResNet-50, ResNet-18, ResNet-34, and ResNet-101 are plotted from attack $ = 0$ to attack $ = 200$. Similar trends were observed when the sample size was increased.}
}
\end{figure}

The first three graphs are polarizing. The model seems to be required to reach a certain depth in order to survive the initial 50 attacks. At 34 layers, the model is nonresistant to the initial attacks, but at 50 layers, the model classifies well on White to Black, Black to White, and Random Pixel Invert attacks. What remains interesting is the gap observed in the Edge to Alter attacks, where depth seems to be much less of a "threshold" needing to be met. Instead, each depth performs notably better than the last, implying that Edge to Alter requires a model with more parameters to test well. 

\section{Conclusion}

We have shown that in general adversarial cases, a greater depth and parameter count in a learner can improve its classification accuracy. We have also demonstrated that non-residual architectures appear to outperform their residual counterparts when standardized for complexity.

Intuition as to why VGG-19 and the FCN performed better than expected can be sought for in the architectural differences, namely that ResNet-50, ResNet-50V2, and EfficientNet-B1 all contain residual shortcut connections. It is possible that ResNet-50's and EfficientNet-B1's worse-than-expected performance is due to their residual connections beginning with a weighted layer, whereas in ResNet-50V2, Batch Normalization and Rectified Linear Unit functions precede the weighted layer. The potential issue of residual skips is absent for both VGG-19 and the FCN, perhaps explaining the cause of their adversarial resistance.

VGG-19 and EfficientNet-B1 do more poorly on Background-to-Foreground (Black to White) attacks compared to Foreground-to-Background (White to Black) attacks, meaning that those models potentially value background information more than foreground information. ResNet-50, ResNet-50V2, and the FCN, however, value background and foreground information nearly equally. This seems counter-intuitive to human convention, where numbers are identified using their foreground image rather than their background components, and it is likely that human eyes will locate the foreground prior to the background.

The most obvious correlation discovered was the effect of model depth on resilience. Due to ResNet's "threshold" values at various depths, an increased model depth and parameter count imply anywhere between a significant increase to no change in classification accuracy. Overall, higher complexity and depth seem to benefit resistance against adversarial attacks, although certain architectural improvements that increase complexity may hinder resistance as well.

\section{Future Work}

We hope to inspire further research on using adversarial attacks to benchmark newly-vetted architectures. In the future, we will likely conduct more rigorous examination of architecture features and their relation to 
model resilience. In particular, we plan to investigate residual networks greatly, experimenting with vast varieties of skip connections between layers to determine their effects on resilience. Along with that, inquiry on model width and image resolution's effects must be established. Future work should include a collection of components (blocks \& layers used, model complexity, etc.) of a Convolutional Neural Network altered independently from one another.

The simplicity of the dataset MNIST cannot be ignored. We wonder if the complexity of MNIST has any ramifications regarding the appearance of a depth "threshold" in classification accuracy. Specifically, we theorize that the threshold disappears when the complexity of the task increases, and that ResNet-101, when classifying more difficult images, will subsequently improve beyond ResNet-50. Further research on this topic is necessary, however, to prove or challenge this hypothesis.

Test set accuracy as a final metric may be useful in determining model performance, but determining \textit{why} the model performs in such a way is more difficult. Additional metrics should be considered, such as a "confidence" metric, capable of being introduced using a simulated Gaussian process\cite{Gal2015DropoutB}. If we assign dropout at random to a network output with input $\mathrm{x*}$ (as if dropouts were being used in training), a sampling distribution of classification accuracies is formed, and its predictive variance $\mathrm{Var(y*)}$ exists as the confidence metric.

We anticipate to move yet another step closer to definitive learner characteristics and metrics that determine learner resilience to perturbations and attacks. This paper offers various angles of insight as to what makes a learner resilient, but more work is needed to determine causation. We hope to have the resources in future undergraduate research necessary to further and solidify our current claims.

{\small
\bibliographystyle{ieee}
\bibliography{egbib}
}

\end{document}